\documentclass[final]{configuration/opt2021} %

\usepackage[T1]{fontenc}
\usepackage[utf8]{inputenc}
\usepackage{enumitem}

\providecommand{\norm}[1]{\left\lVert#1\right\rVert}

\providecommand{\EEb}[2]{{\mathbb E}_{#1}\left[#2\right] } %

\renewcommand{\gg}{\mathbf{g}}

\providecommand{\mm}{\mathbf{m}}

\providecommand{\vv}{\mathbf{v}}

\providecommand{\xx}{\mathbf{x}}
\providecommand{\yy}{\mathbf{y}}
\providecommand{\zz}{\mathbf{z}}

\providecommand{\cD}{\mathcal{D}}

\providecommand{\cI}{\mathcal{I}}

\providecommand{\cL}{\mathcal{L}}

\providecommand{\cX}{\mathcal{X}}
\providecommand{\cY}{\mathcal{Y}}

\newenvironment{talign*}
{\csname align*\endcsname}
{\endalign}

\title[Understanding Memorization]{Understanding Memorization from the Perspective of Optimization via Efficient Influence Estimation}

\optauthor{%
\Name{Futong Liu} \Email{futong.liu@epfl.ch}\\
\Name{Tao Lin}\thanks{Corresponding author} \Email{tao.lin@epfl.ch}\\
\Name{Martin Jaggi} \Email{martin.jaggi@epfl.ch}\\
\addr EPFL, Switzerland
}

\begin{document}

\maketitle

\begin{abstract}%
	Over-parameterized deep neural networks are able to achieve excellent training accuracy while maintaining a small generalization error. It has also been found that they are able to fit arbitrary labels, and this behaviour is referred to as the phenomenon of memorization.
	In this work, we study the phenomenon of memorization with turn-over dropout, an efficient method to estimate influence and memorization, for data with true labels (real data) and data with random labels (random data).
	Our main findings are: (i) For both real data and random data, the optimization of easy examples (e.g., real data) and difficult examples (e.g., random data) are conducted by the network simultaneously, with easy ones at a higher speed; (ii) For real data, a correct difficult example in the training dataset is more informative than an easy one.
	By showing the existence of memorization on random data and real data, we highlight the consistency between them regarding optimization and we emphasize the implication of memorization during optimization.
\end{abstract}

\setlength{\parskip}{-2pt plus -2pt minus 2pt}

\section{Introduction}
Over-parameterized deep neural networks (DNNs), which have more number of trainable parameters than that of data instances, usually achieve excellent performance on both the training dataset and the test dataset.
However, it has also been found that a DNN has the capability to perfectly fit pure random labels, and this behaviour of a DNN when trained on random labels or random inputs has been implicitly referred to as the phenomenon of memorization \citep{zhang2017understanding,arpit2017closer,brown2021memorization,maennel2020neural}.
Understanding the memorization of deep neural networks is a key step towards characterizing its optimization dynamics and interpreting the model's generalization ability.

\citet{zhang2017understanding} formulated the memorization phenomenon by showing that a neural network with sufficient capacity can memorize completely random labels or random noise data (inputs) without substantially longer training time, and the explicit regularization is neither necessary nor sufficient to control the generalization error.
\citet{arpit2017closer} further varied the amount of random labels, and concluded that DNNs first learn general patterns of the real data before fitting the random data.

The phenomenon of memorization can also be observed on a real dataset when there are no random labels.
\citet{feldman2020neural} studied memorization through the lens of self-influence estimation, and provided compelling evidences for the long tail theory \citep{feldman2020does}. They believe that memorization is necessary for achieving close-to-optimal generalization error when the data distribution is long-tailed.
Modern datasets usually follows a long tail distribution, where the subpopulations at the head of the distribution include a large amount of frequent and easy examples while the subpopulations at the tail include rare and difficult examples.

However, the insights of these prior work normally are limited through some computationally intensive estimation.
For example,~\citet{arpit2017closer, gu2019neural} only study the memorization phenomenon on random data, and it is not clear if such observations can be generalized to real data.
Similarly, the study on influence estimation \citep{feldman2020neural, koh2017understanding} focuses on the cross-influence of a training instance over a test instance, instead of self-influence nor its implication on optimization.
To this end, we introduce the memorization score (i.e.\ self-influence)---as a unified and computationally efficient metric---on both real data and random data to estimate the difficulty of training instances; such efficient estimation allows us to study the interpolation between memorization, optimization, and its implication on generalization.

Our main contributions are summarized as follows:
\begin{enumerate}[nosep,leftmargin=12pt]
	\item We leverage turn-over dropout to quantify the degree of memorization by estimating its memorization score.
	      We show it is feasible and efficient on different network architectures.
	\item For real data and random data, easy and difficult examples are learnt simultaneously.
	      However, the optimization on easy examples converges faster than that of difficult ones, due to the higher gradient similarity between easy examples.
	\item For real data, difficult examples contain information about easy ones but easy examples have very little information about difficult ones. Easy examples can benefit from both easy and difficult examples, and hence the optimization speed of easy examples is faster.
\end{enumerate}

\section{Influence Estimation with Turn-over Dropout}
\subsection{Preliminary}

We denote $\cD = \{\xx_i, \yy_i\}_{i=1}^N$ as the training dataset, $\cX$ as the input space (e.g., images) and $\cY$ as output space (e.g., labels).
Let $\zz_i:=(\xx_i, \yy_i)$, where $\zz$ is a data example and represent a pair of input $\xx_i \in \cX$ and its output $\yy_i \in \cY$.
A model trained on dataset $\cD$ is denoted as $f_\cD: \cX \rightarrow \cY $.
The loss function $\cL(f, \zz_i)$ represents the loss of the model's prediction on $\xx_i$ over its label $\yy_i$.

The phenomenon of memorization and generalization can be quantified by the \textbf{leave-one-out influence} function \citep{feldman2020neural,koh2017understanding}, which measures the influence of example $\zz_i$ to the prediction of $\zz_{target}$.
Let $f_{\cD}$ be the model trained on the full training dataset $\cD$ and $f_{\cD \backslash \{\zz_{i}\}}$ be the model trained on $\cD$ excluding example $\zz_i$, then the influence is defined as:
\begin{equation} \label{eq: inter-influence}
	\cI(\zz_{target}, \zz_{i} ; \cD) :=
	\cL(f_{\cD \backslash \{\zz_{i}\}}, \zz_{target})- \cL(f_{\cD}, \zz_{target}) \,,
\end{equation}
where $\zz_{target}$ can be an instance in the training set or the test set.
When $\zz_{target} = \zz_{i} \in \cD$, \textbf{memorization} can be defined as \textbf{self-influence} \citep{feldman2020neural}:
\begin{equation} \label{eq: self-influence}
	\cI( \zz_{i}, \zz_{i} ; \cD) :=
	\cL( f_{ \cD \backslash \{ \zz_{i}\}}, \zz_{i}) - \cL(f_{\cD}, \zz_{i}) \,.
\end{equation}
The memorization score computed by Equation \eqref{eq: self-influence} indicates the extent of memorization.
A large memorization score means that $\zz_i$ is memorized to be fitted, as the model that is trained without $\zz_i$ cannot correctly classify $\zz_i$.
A small score indicates that the label of $\zz_i$ can be inferred from other training data, and hence $\zz_i$ is part of a generalizable pattern learnt by the model.

For a more fine-grained understanding, we define the \textbf{cosine similarity} of a set of gradient vectors $\{v_i\}, i \in [n]$ as the average pair-wise cosine similarities, as
\begin{align}
	\text{cosine similarity} = \EEb{i, j}{
	\frac{ \vv_{i} \cdot \vv_{j} }{ \norm{\vv_i}_2 \cdot \norm{\vv_j}_2 }
	}, \;\; \forall i, j \in [n] \: and \: i \neq j \,.
	\label{eq: cosine sim}
\end{align}
We further define the \textbf{contribution} of an example's gradient to the actual gradient update step as the projection of the example's gradient on the gradient of the mini-batch $\gg$, i.e.\
\begin{align}
	\text{contribution of } \vv_i =
	(\vv_{i} \cdot \gg) / \norm{\gg}_2
	\,.
	\label{eq: contribution}
\end{align}

\subsection{Turn-over Dropout as an Efficient Estimation for Memorization}
It is computationally infeasible to calculate the exact leave-one-out influence and memorization scores through Equation \eqref{eq: inter-influence} and \eqref{eq: self-influence}, due to the significant overhead introduced by training the extra model from scratch for each training instance.

\begin{figure}[tb]
	\centering
	\includegraphics[width=0.6\columnwidth]{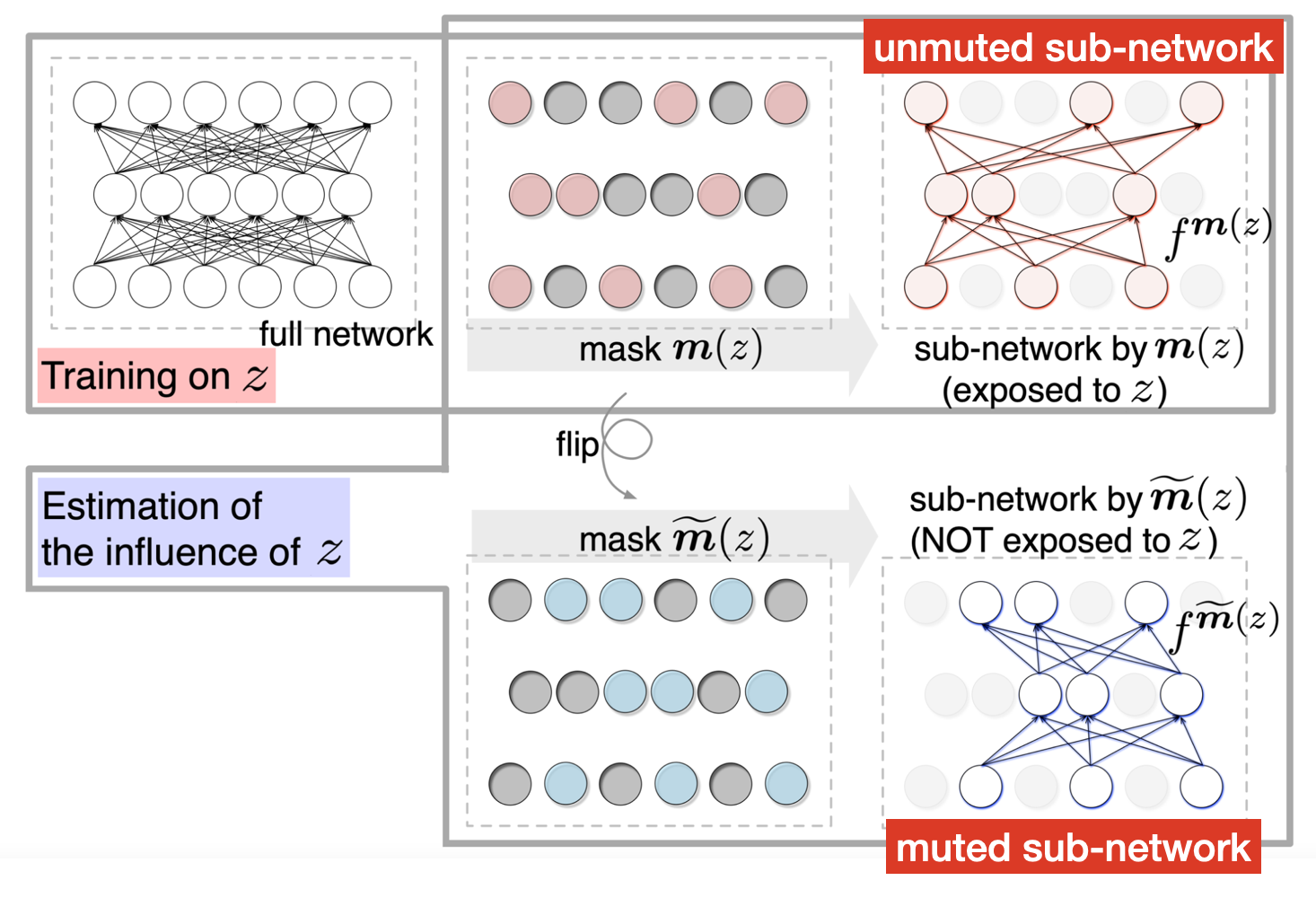}
	\vspace{-2em}
	\caption{\small
		\textbf{The turn-over dropout mechanism}.
		Turn-over dropout generates a deterministic mask for each training example $z$ and only updates the un-muted sub-network.
		The muted sub-network is not updated with respect to $z$.
		The influence estimation is the loss difference of the two sub-networks.
		Cited and adapted from \cite{kobayashi2020efficient}.
	}
	\vspace{-2em}
	\label{fig:turn_over_dropout}
\end{figure}

For the ease of getting a deeper understanding on the optimization and generalization (as our main contribution), we follow the idea of \textbf{turn-over dropout}~\citep{kobayashi2020efficient}, which is stated as follows (and the workflow refers to Figure \ref{fig:turn_over_dropout}).
We also visualize some cases of easy and difficult examples identified by turn-over dropout in Appendix~\ref{Easy and Difficult Examples Identified by Turn-over Dropout}.

The turn-over dropout generates a \emph{deterministic} mask $\mm(\zz_i)$ for each training instance $\zz_i$, and the mask is fixed throughout training and evaluation. The masks are controlled to ensure no two training instances will share the same mask, and each mask splits the model into two sub-networks: the muted sub-network and the unmuted sub-network.

We use $f^{\tilde{\mm}(\zz_i)}$ and $f^{\mm(\zz_i)}$ to approximate $f_{\cD \backslash \{\zz_{i}\}}$ and $f_{\cD}$ respectively, given the fact that the muted sub-network $f^{\tilde{\mm}(\zz_i)}$ is updated by $\cD \backslash \{\zz_{i}\} $ and the un-muted sub-network $f^{\mm(\zz_i)}$ is updated by $\cD$.
Hence, the influence score in Equation~\eqref{eq: inter-influence} can be estimated by
\begin{equation} \label{eq: turn-over inter-influence}
	\cI(\zz_{target}, \zz_{i} ; \cD) \approx
	\cL(f^{\tilde{\mm}(\zz_i)}, \zz_{target})- \cL(f^{\mm( \zz_i )}, \zz_{target}) \,,
\end{equation}
and the memorization (self-influence) score can be estimated by
\begin{equation} \label{eq: turn-over self-influence}
	\cI(\zz_{i}, \zz_{i} ; \cD) \approx
	\cL(f^{\tilde{\mm}(\zz_i)}, \zz_{i})- \cL(f^{\mm( \zz_i)}, \zz_{i}) \,.
\end{equation}

Note that the prior work~\citep{kobayashi2020efficient} uses turn-over dropout as an addition to DNNs to increase the model's interpretability.
With turn-over dropout only added to deep layers, their implementation does not sacrifice the model's performance too much.
Our work instead applies turn-over dropout to all layers and uses the adapted model as a proxy to estimate self-influence.
Furthermore, we extend the self-influence information to the training procedure and study its implication on optimization.

\section{Observations and Insights}

\subsection{Unify random data and real data by memorization scores}
\label{Unify random data and real data by memorization scores}

A commonly-used method to study memorization is to \emph{corrupt} the training data by replacing part of true labels with random labels.
With this method, previous work \citep{zhang2017understanding,arpit2017closer,maennel2020neural} focuses on comparing the \emph{differences} between the scenario of label corruption and the scenario of no corruption to understand memorization.
On the contrary, we juxtapose the two scenarios and focus on comparing the easy and difficult examples to find what is in \emph{common} regarding optimization.

\begin{figure}[!b]
	\centering
	\vspace{-1em}
	\subfigure[\small Uncorrupted MNIST.]{
		\label{fig:mnist memorization scores a}
		\includegraphics[width=0.475\textwidth]{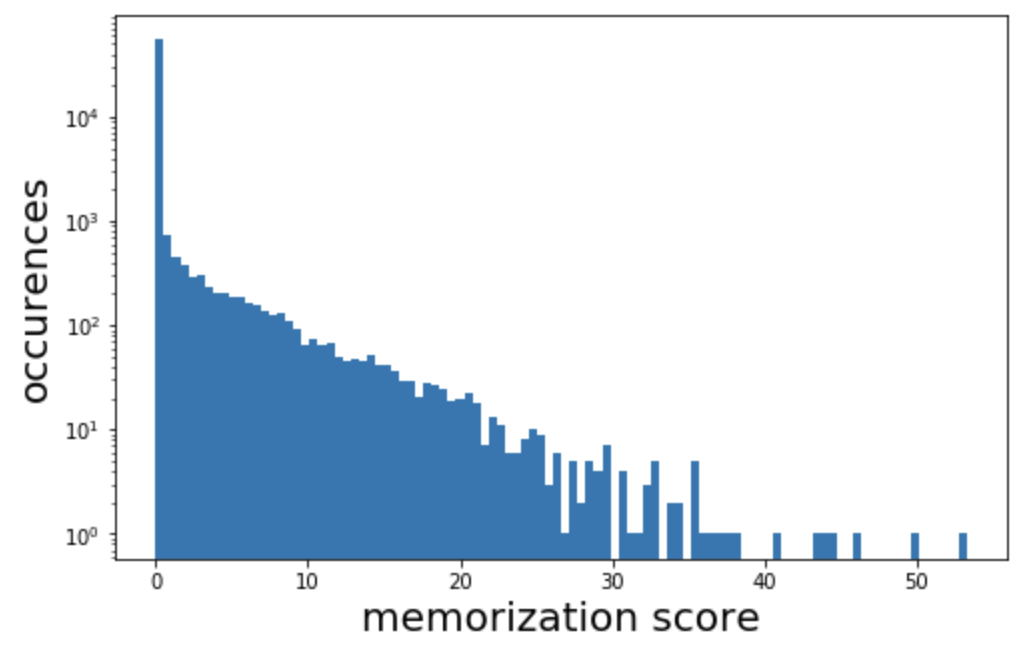}
	}
	\hfill
	\subfigure[\small MNIST corrupted with 20,000 random labels.]{
		\label{fig:mnist memorization scores b}
		\includegraphics[width=0.475\textwidth]{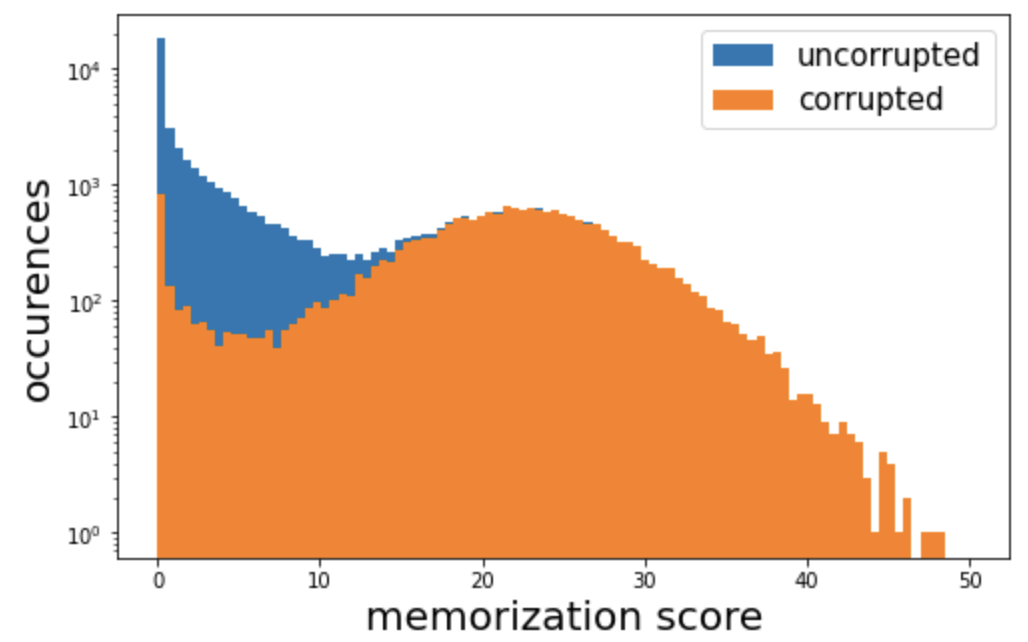}
	}
	\vspace{-1em}
	\caption{\small
		\textbf{Memorization scores of MNIST} computed from a one-hidden layer MLP with turn-over dropout. (a) Data follows a long tail distribution. (b) Data with random labels (corrupted) have higher memorization scores than data with real labels (uncorrupted).
	}
	\vspace{-2em}
	\label{fig:mnist memorization scores}
\end{figure}

A corrupted dataset is a more difficult version of the uncorrupted one.
Figure \ref{fig:mnist memorization scores} shows the histogram of the memorization scores (computed from turn-over dropout) on uncorrupted MNIST and corrupted MNIST.
When there is no corruption, the scores follow a long tail distribution as described in \citet{feldman2020does} (c.f.\ Figure~\ref{fig:mnist memorization scores a}).
Corrupting an instance significantly increases its difficulty (memorization score): due to the random pattern/relationship between the image and the label, the model has no option but to \emph{memorize} the instance.
Random examples in the corrupted scenario are the counterpart of the difficult examples in the uncorrupted scenario, as shown in Figure~\ref{fig:mnist memorization scores b}.
Hence, \emph{random data and real data are consistent with respect to memorization scores}, and the methodology to study the properties of random data sheds light on difficult examples in the uncorrupted scenario.

\subsection{Training on random labels} \label{sec:random_labels}
In this section, our investigation starts from the revisiting on the experiments of \citet{arpit2017closer} by corrupting the dataset with different degree of corruption.

\paragraph{Easy examples are learnt faster.}
In contrast to the common belief introduced in \citet{arpit2017closer}\footnote{
	We also reproduce these experiments in Appendix \ref{training_dynamics_on_random_labels}.
} that a model first learns the simple and general patterns (easy examples) before fitting the noise (difficult examples), we provide a refined insight below: \emph{easy and difficult examples are learnt by the model simultaneously, with easy examples learnt faster}.
As indicated in Section~\ref{Unify random data and real data by memorization scores}, in the presence of label corruption, data with random labels are of the highest difficulty.

\begin{figure}[!h]
	\centering
	\vspace{-0.5em}
	\subfigure[\small Training accuracy.]{
		\label{fig:mnist 20k training acc}
		\includegraphics[width=0.475\textwidth]{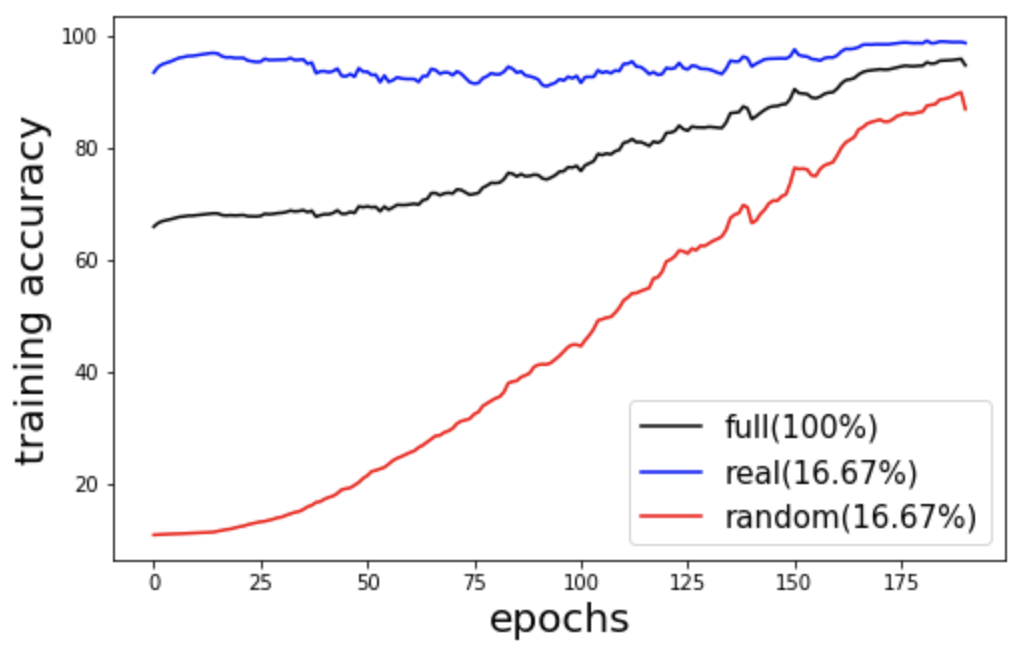}
	}
	\hfill
	\subfigure[\small Test accuracy.]{
		\label{fig:mnist 20k test acc}
		\includegraphics[width=0.475\textwidth]{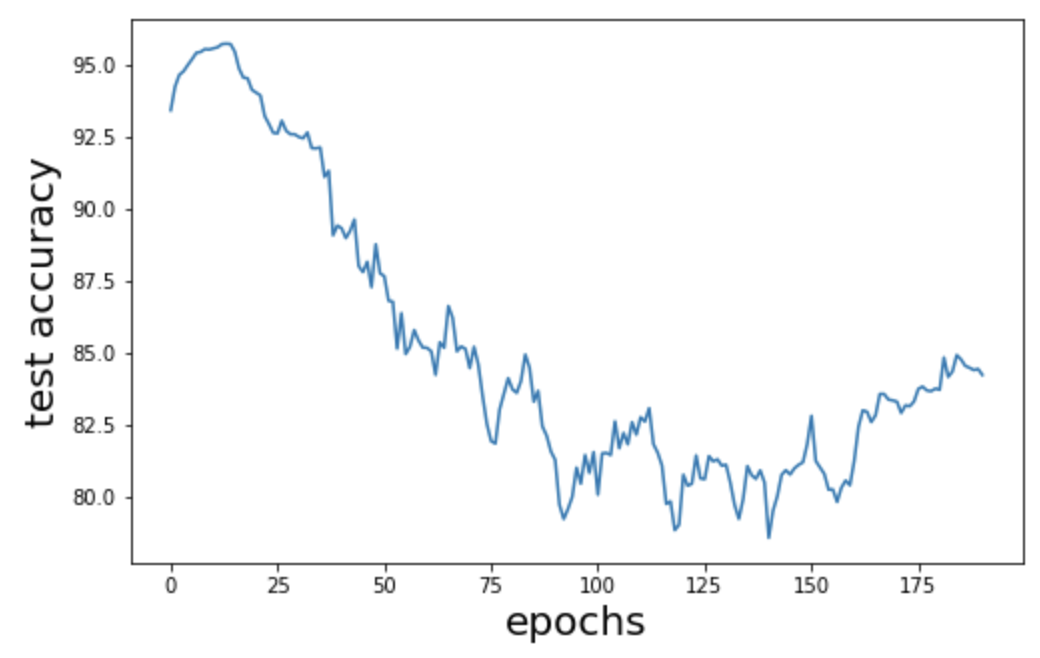}
	}
	\vspace{-1em}
	\caption{\small
		\textbf{The training and test dynamics} of a one-hidden-layer \textbf{MLP on the MNIST} dataset when 20,000 of the training dataset is corrupted by random labels. Results are sliding-window smoothed.
		The black line (``full'') in (a) and the line in (b) shows the dynamics on the entire training/test dataset respectively.
		(a) also keeps track of the training accuracy of a 10,000 (16.67\%) real-data subset (blue) and a 10,000 (16.67\%) random-label subset (red). The percentage in the legend indicates the size of the subset.
		(b) shows that the test accuracy reaches the peak with the build-up of the speed difference and drops with the random data improving its training accuracy.
	}
	\vspace{-2em}
	\label{fig:mnist 20k training and test acc}
\end{figure}

Figure \ref{fig:mnist 20k training and test acc} shows the training and test dynamics of a one-hidden-layer MLP on the MNIST dataset with 20,000 instances replaced by random labels.
We keep track of training dynamics of an easy subset and a difficult subset of the training data.
It can be observed that easy examples are learnt much faster than difficult ones. The build-up of the speed difference reflects on the dynamic of test accuracy: the test accuracy reaches the peak (at around 20th epoch), and that is when the model has successfully fitted the real data but barely fitted the random noise yet.

\begin{figure}[!t]
	\centering
	\subfigure[\small Cosine similarity]{
		\includegraphics[width=0.475\textwidth]{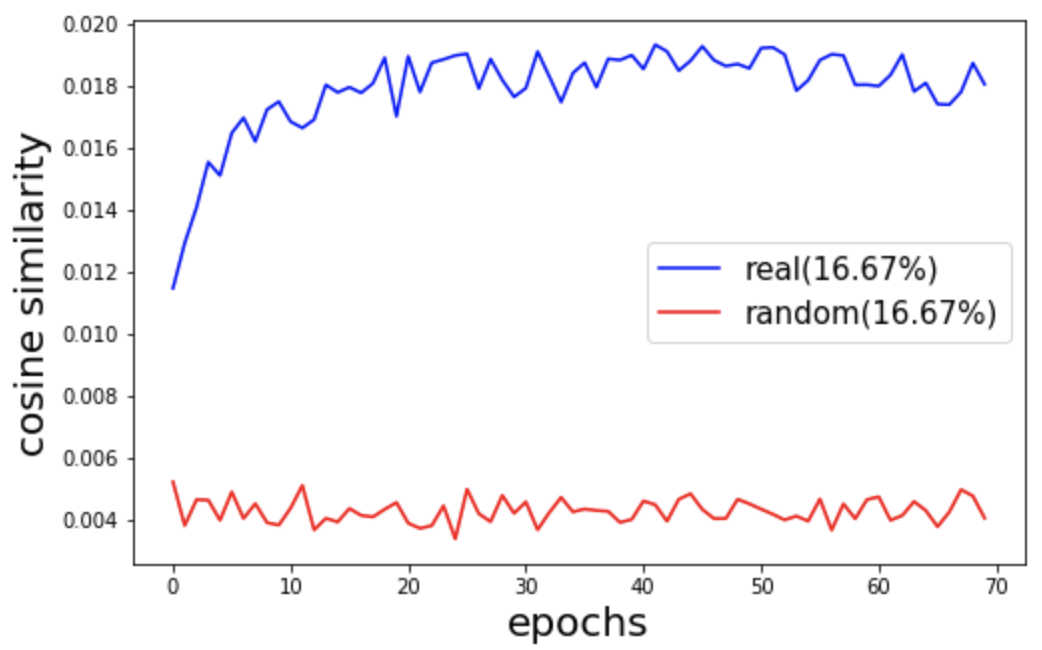}
		\label{fig:mnist_20k_random_cosine_similarity}
	}
	\hfill
	\subfigure[\small Contribution]{
		\label{fig:mnist 20k contribution}
		\includegraphics[width=0.475\textwidth]{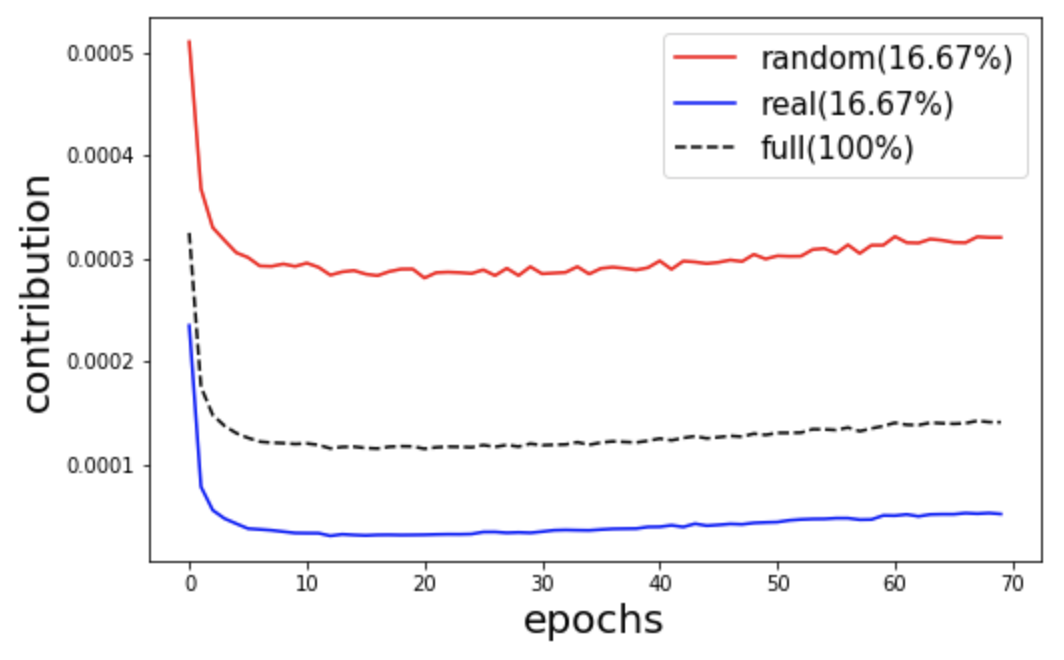}
		\label{fig:mnist_20k_random_contributions}
	}
	\vspace{-1em}
	\caption{\small
		Dynamics of \textbf{cosine similarity} and \textbf{contribution} of an easy (real) subset and a difficult (random) subset of the training data (\textbf{MLP} trained on \textbf{MNIST} with 20,000 instances corrupted). (a) The cosine similarity between easy examples is much higher. (b) The difficult examples are the driving force of SGD update throughout training.
	}
	\vspace{-2em}
	\label{fig:mnist 20k cosine similarity and contribution}
\end{figure}

However, we also observe from Figure \ref{fig:mnist 20k training acc} (and Figure \ref{fig:sectioned dynamics on mnist and cifar} later) that there are small improvement on the difficult subset at the initial epochs. Albeit the improvement is small, it indicates that difficult examples are in fact learnt simultaneously with easy examples, and Figure \ref{fig:mnist 20k contribution} in the next section shows stronger proof on this point.

\paragraph{Why are easy examples learnt faster than difficult examples?}
Below we try to answer the question from the perspective of optimization (SGD updates) of real data and random data\footnote{
	We use Backpack~\citep{dangel2019backpack} to extract the gradient of each example in the batch during the backward pass.
}.
Figure \ref{fig:mnist 20k cosine similarity and contribution} shows the evolution of the average cosine similarity (Equation \eqref{eq: cosine sim}) and the average contribution (Equation \eqref{eq: contribution}) of easy examples and difficult examples.

The cosine similarity between easy examples is much higher than that of difficult ones, as shown in Figure~\ref{fig:mnist_20k_random_cosine_similarity}.
Easy examples reside at the head of the long tail distribution and belong to similar subpopulations, and this similarity between images engenders similar gradients. These similar gradients strengthen each other to make a SGD update that can benefit them all, which results in the faster optimization speed for easy examples.
This observation is in agreement with \cite{chatterjee2020coherent}, who uses a measure of gradient coherence to show that the parameter update of the network can benefit many examples when such similarity exists.

Additionally, we observe from Figure \ref{fig:mnist 20k contribution} that the average contribution of difficult examples is constantly higher than easy examples and the entire dataset.
The high contribution of difficult examples throughout training shows that difficult examples are learnt simultaneously with easy examples, in contrast to the prior belief that DNNs prioritize the learning of easy examples.

\subsection{Training on real data}
\begin{figure}[!h]
	\centering
	\subfigure[\small Training accuracy on MNIST.]{
		\includegraphics[width=0.47\textwidth]{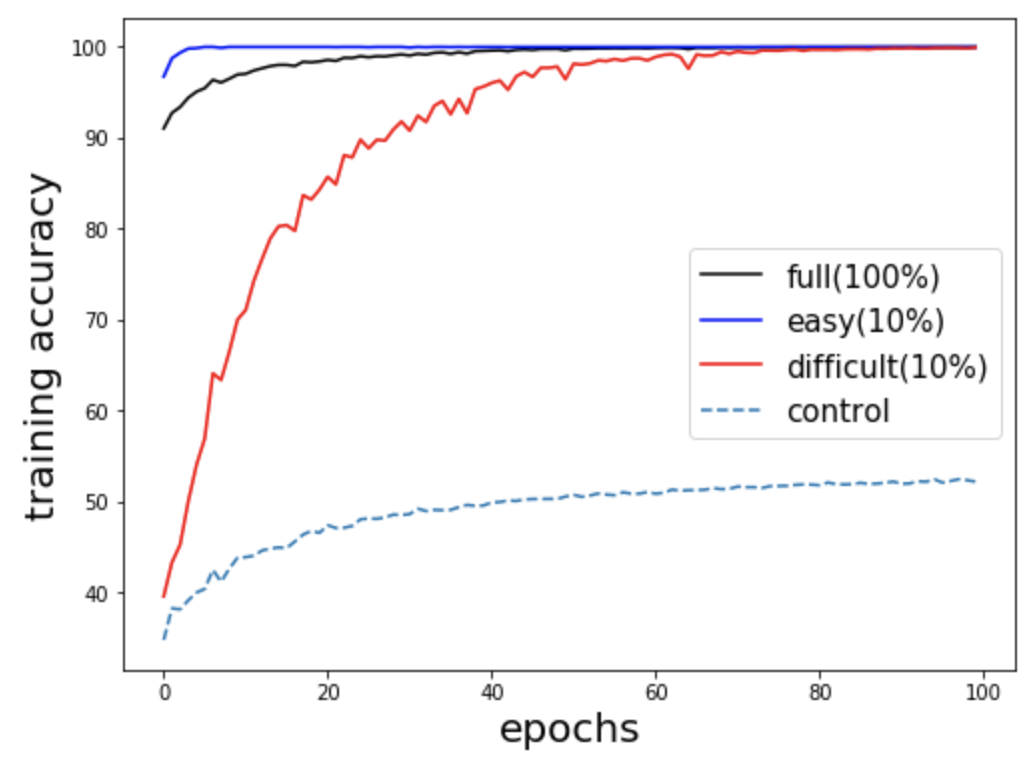}
		\label{fig:mnist_easy_and_difficult_learnt_at_the_same_time_a}
	}
	\hfill
	\subfigure[\small Test accuracy on MNIST.]{
		\includegraphics[width=0.47\textwidth]{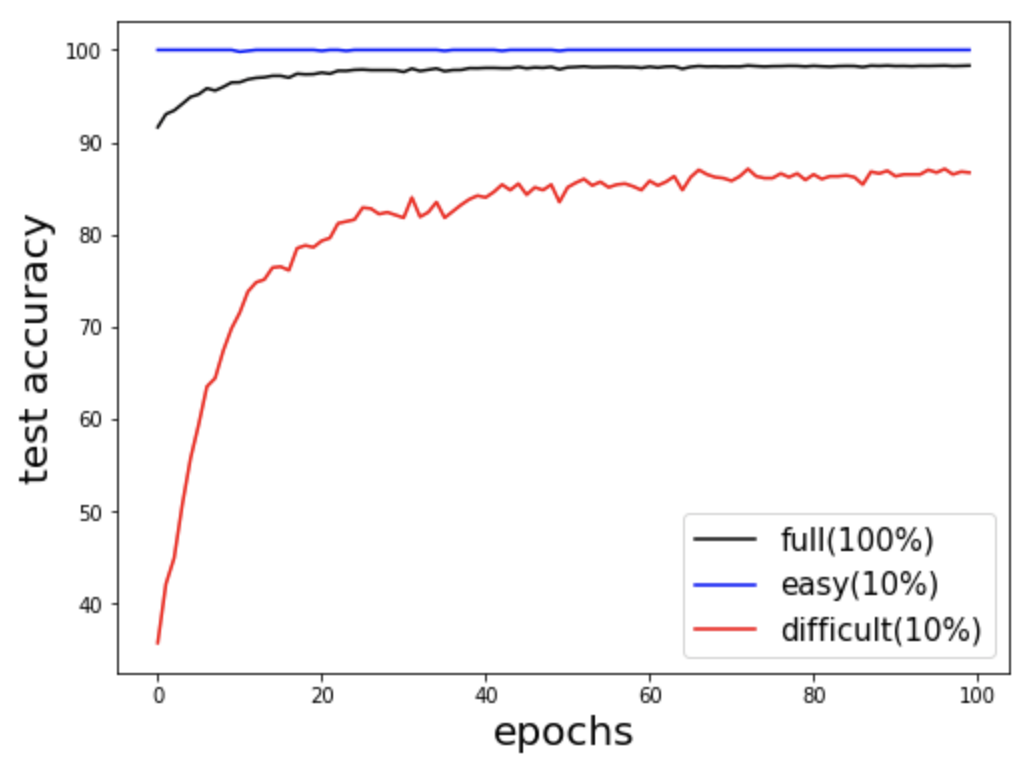}
		\label{fig:mnist_easy_and_difficult_learnt_at_the_same_time_b}
	}
	\vfill
	\subfigure[\small Training accuracy on CIFAR-10.]{
		\includegraphics[width=0.47\textwidth]{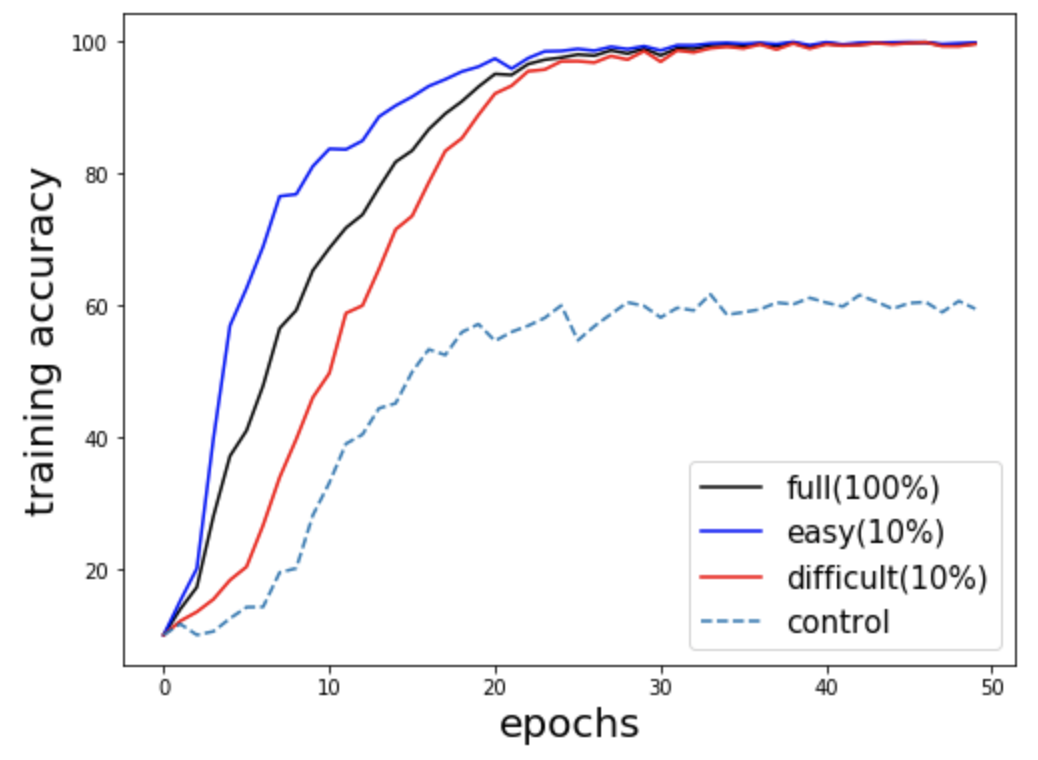}
		\label{fig:cifar10_easy_and_difficult_learnt_at_the_same_time_a}
	}
	\hfill
	\subfigure[\small Test accuracy, CIFAR-10]{
		\includegraphics[width=0.47\textwidth]{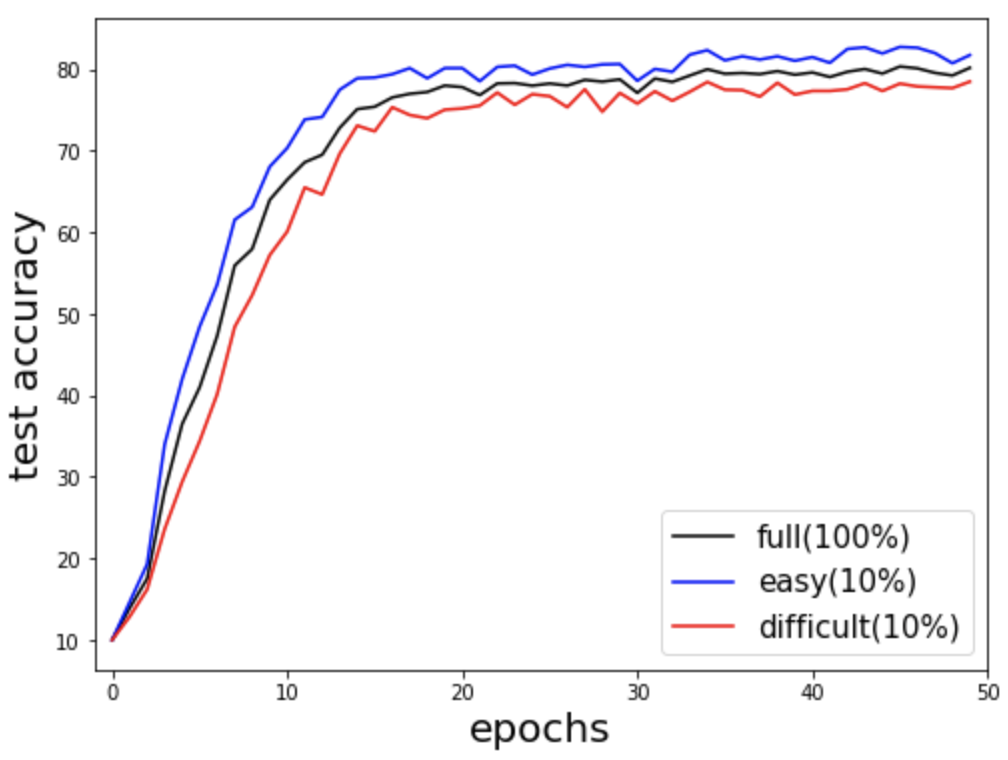}
		\label{fig:cifar10_easy_and_difficult_learnt_at_the_same_time_b}
	}
	\vspace{-1em}
	\caption{\small
		\textbf{Training and test dynamics on easy and difficult subsets} of MNIST (with MLP) and CIFAR-10 (with VGG-11). The memorization scores are used to partition the dataset and the percentage in parentheses indicates the size of the subset. Difficult examples (red) is learnt simultaneously with easy examples. As a control, the dotted line in (a) and (b) indicates training on the training data with the difficult subset removed. Thus, we confirm that the improvement of difficult examples' training accuracy is not a side effect of the patterns learnt from easy examples.
	}
	\vspace{-2em}
	\label{fig:sectioned dynamics on mnist and cifar}
\end{figure}

\paragraph{Training dynamics on real data.}
Figure \ref{fig:sectioned dynamics on mnist and cifar} shows how the easy/difficult/full subsets of the training/test data evolve with respect to epochs on MNIST and CIFAR-10.
We compute the memorization scores of the training data and test data beforehand, and monitor an easy subset and a difficult subset during training and validation.
The observation is consistent with the random-label case: \emph{easy examples and difficult examples are learnt simultaneously, with easy examples fitted faster than difficult ones.}

Figure \ref{fig:mnist_easy_and_difficult_learnt_at_the_same_time_b} and \ref{fig:cifar10_easy_and_difficult_learnt_at_the_same_time_b} shows the test dynamics are similar to the training dynamics: \emph{easy examples benefit from learning quicker than difficult ones and reach a higher test accuracy}. This similarity reflects the consistency of the subpopulations identified.

\begin{figure}[!t]
	\centering
	\subfigure[\small Train on an easy subset.]{
		\includegraphics[width=0.475\textwidth]{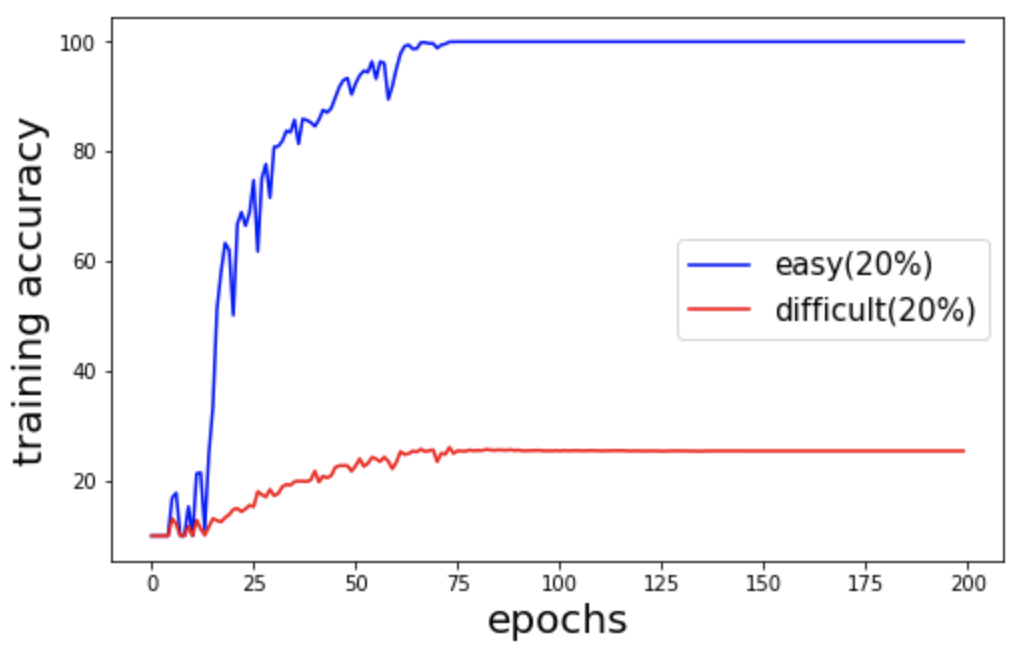}
		\label{fig:cifar10-train_on_easy_training_acc}
	}
	\hfill
	\subfigure[\small Train on a difficult subset.]{
		\includegraphics[width=0.475\textwidth]{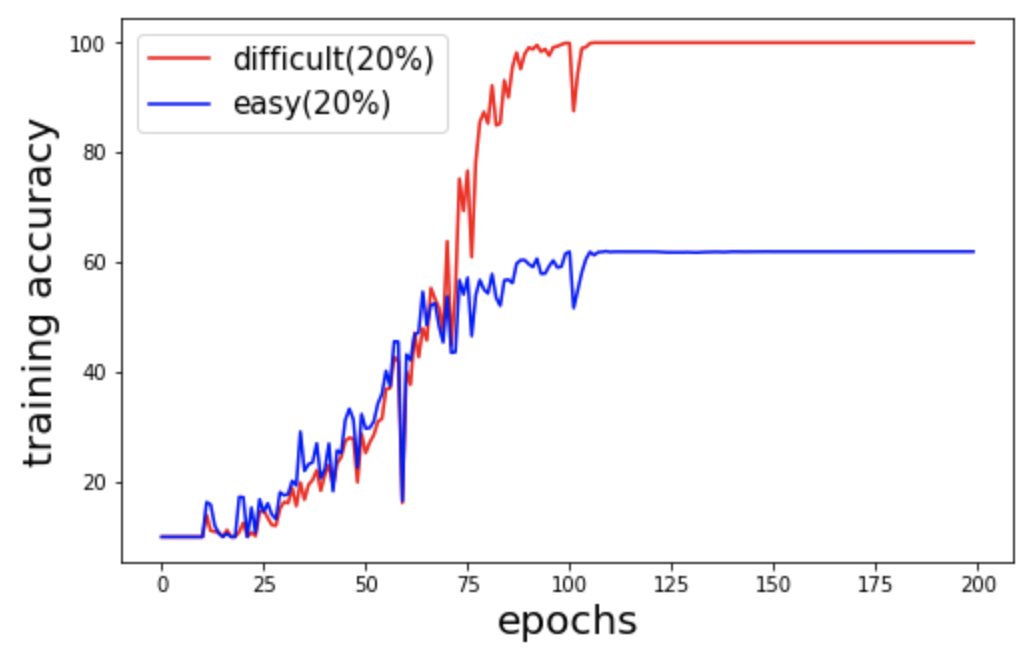}
		\label{fig:cifar10-train_on_difficult_training_acc}
	}
	\vspace{-1em}
	\caption{\small
		\textbf{Training dynamics on two disjoint subsets of CIFAR-10 with VGG-11}.  (a) shows the training accuracy when trained only with the easy subset. (b) shows the training accuracy when trained only with the difficult subset. Difficult examples contain information about the easy ones while the easy examples has very little information about the difficult ones.
	}
	\vspace{-2em}
	\label{fig:difficult examples contains info about easy ones}
\end{figure}

\paragraph{Difficult examples are more informative.}
As explained in the random label setting (Section~\ref{sec:random_labels}), easy examples are learnt faster than difficult ones due to the higher gradient similarity of easy examples.
Complementing this observation, we further highlight a new insight in Figure \ref{fig:difficult examples contains info about easy ones} below for real data, which again explains \emph{the faster convergence on the easy examples}, that is \emph{difficult examples contain information about the easy ones while the easy examples have very little information about the difficult ones}.

Figure \ref{fig:difficult examples contains info about easy ones} presents the training dynamics of two manually-constructed sub-datasets: with the memorization scores computed for the training data, we sample an easy subset and a difficult subset that are of the same size from the whole training dataset; and for each time, we train with one subset, and use the other one as the validation set.
We can witness from Figure~\ref{fig:cifar10-train_on_easy_training_acc} that training on the easy subset enables a fast convergence on the corresponding training accuracy, but a marginal improvement on that of the difficult subset.
While in Figure~\ref{fig:cifar10-train_on_difficult_training_acc}, despite the relatively slow convergence on the difficult subset, the training benefits both the difficult subset and the easy subset substantially.
It proves that the memorization of difficult examples improves the model's ability to generalize on easy examples, while the generalizable patterns learnt from easy examples do not transfer to difficult examples.

A difficult but correct example contains some (relatively weak) degree of similarity with the easy examples, and it results in a slow but still increasing training accuracy on the easy subset when the model is trained on the difficult subset.
A difficult subset contains more (atypical and exclusive) subpopulations than an easy subset and thus is more \textbf{diverse}, so a model that is trained on an easy subset is very unlikely to acquire this unique information.
Therefore, we conclude that difficult examples are more informative and to memorize difficult examples accelerates and benefits the model's ability to generalize.

\section{Conclusion}
Our work studies memorization regarding optimization in the context of random data and real data. We use turn-over dropout to efficiently evaluate memorization scores of training examples, and we unify random data with real data by showing that examples with random labels are simply extremely difficult examples.
We find there are qualitative differences between the easy and difficult examples, and it is consistent for random data and real data that easy and difficult examples are learnt simultaneously, with difficult ones learnt slower.
On real data, we also conclude that difficult examples contain more information than easy ones, and to memorize difficult examples may improve the model's ability to generalize, which may inspire core-set selection or more efficient training algorithms as future work.

\section*{Acknowledgements}
We acknowledge funding from a Google Focused Research Award, Facebook, and European Horizon 2020 FET Proactive Project DIGIPREDICT.

\clearpage
\bibliography{paper}

\clearpage
\appendix

\section{Experiment Set-up}
For MNIST (\cite{mnist_dataset}), we use a one-hidden layer MLP with 4096 neurons in the hidden layer trained with a 0.06 learning rate on 200 epochs and no momentum, except for Figure \ref{fig:mnist 20k training and test acc}, in which we switch to a learning rate of 0.01 for clearer result. 
For CIFAR-10 (\cite{cifar10_dataset}), we use a VGG-11 network (\cite{vgg}) without batch-norm, and it is trained on 200 epochs, with the learning rate starts at 0.01 (with 0.9 momentum) and divided by 10 at epoch 100 and epoch 150 respectively. Additionally, spatial turn-over dropout is applied to VGG-11 on its convolutional layers. The batch size is 256 for training and all the models use the ReLU activation. 

\section{Implementation of Turn-over Dropout}
For linear layers, turn-over dropout can be efficiently implemented to mask the neurons for two sub-networks.
For convolutional layers, 2D spatial turn-over dropout is implemented, which deterministically masks filters (feature maps) instead of neurons.
Half of the feature maps will be set to zero, and the other half will be passed to the next layer.

\section{Easy and Difficult Examples Identified by Turn-over Dropout}
\label{Easy and Difficult Examples Identified by Turn-over Dropout}
\begin{figure}[h]
	\centering
	\subfigure[Low memorization score examples on MNIST dataset.]{\includegraphics[width=0.96\textwidth]{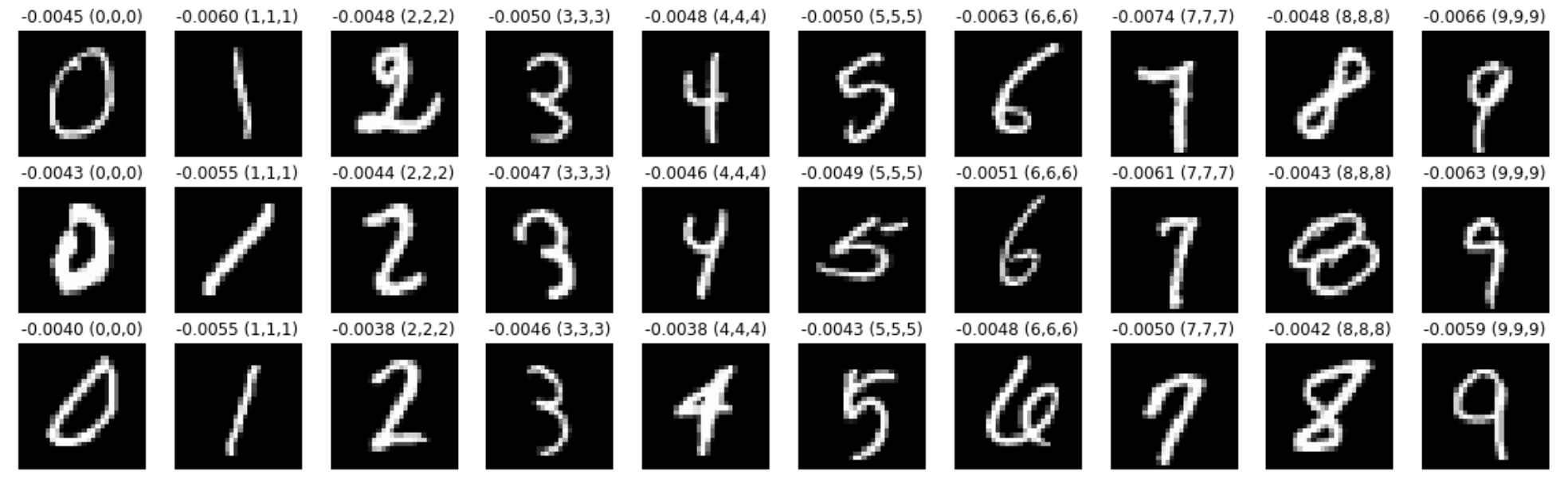}}%
	\vfill
	\subfigure[High memorization score examples on MNIST dataset.]{\includegraphics[width=0.96\textwidth]{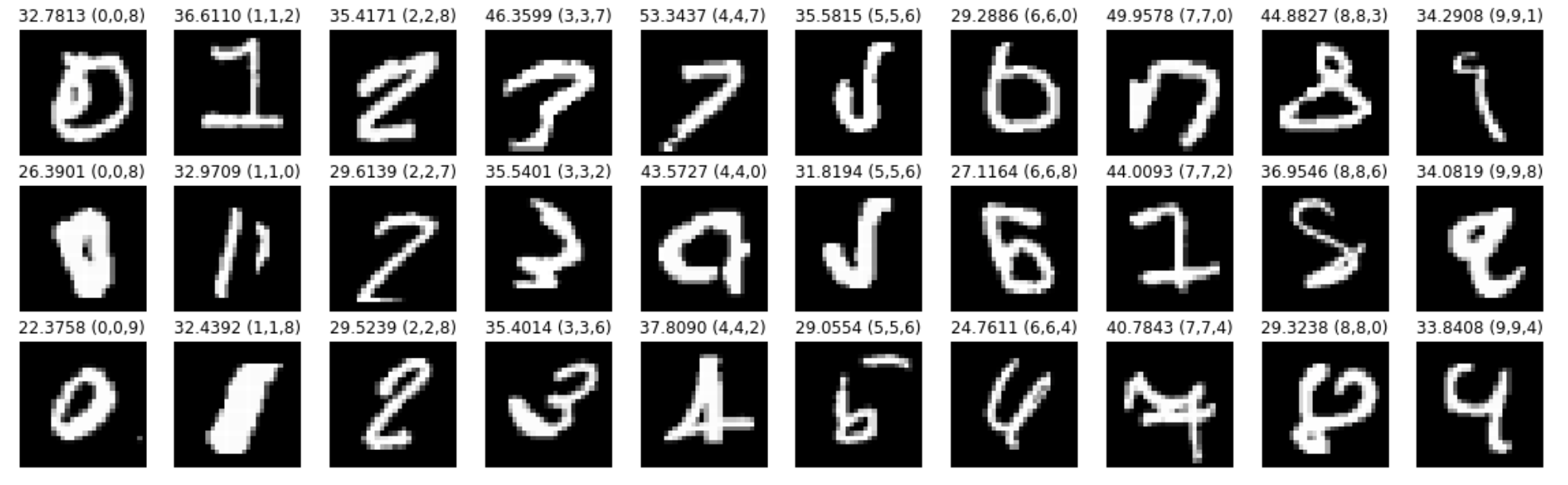}}%
	\caption{Low and high memorization score examples on MNIST dataset. Each column corresponds to one class.
		The number above the image represents the memorization score and the triplet above the image represents (ground truth label, unmuted sub-network's prediction, muted sub-network's prediction).}
	\label{fig:mnist easy and difficult examples}
\end{figure}

\begin{figure}[h]
	\centering
	\subfigure[Low memorization score examples on CIFAR-10 dataset.]{\includegraphics[width=0.96\textwidth]{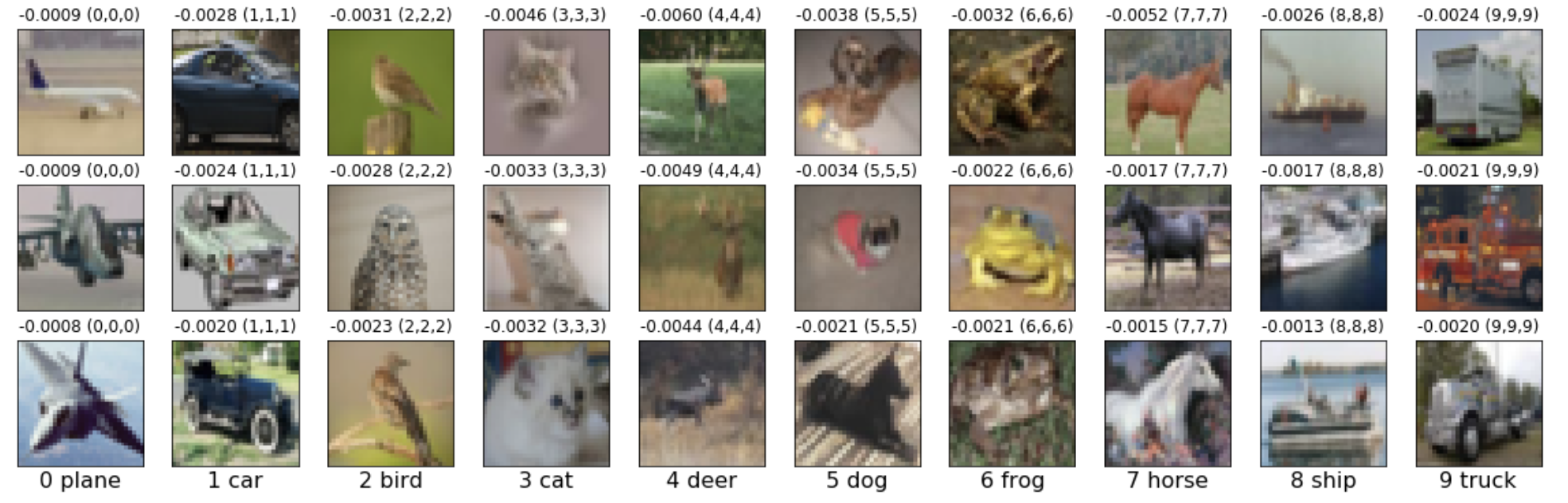}}%
	\vfill
	\subfigure[High memorization score examples on CIFAR-10 dataset.]{\includegraphics[width=0.96\textwidth]{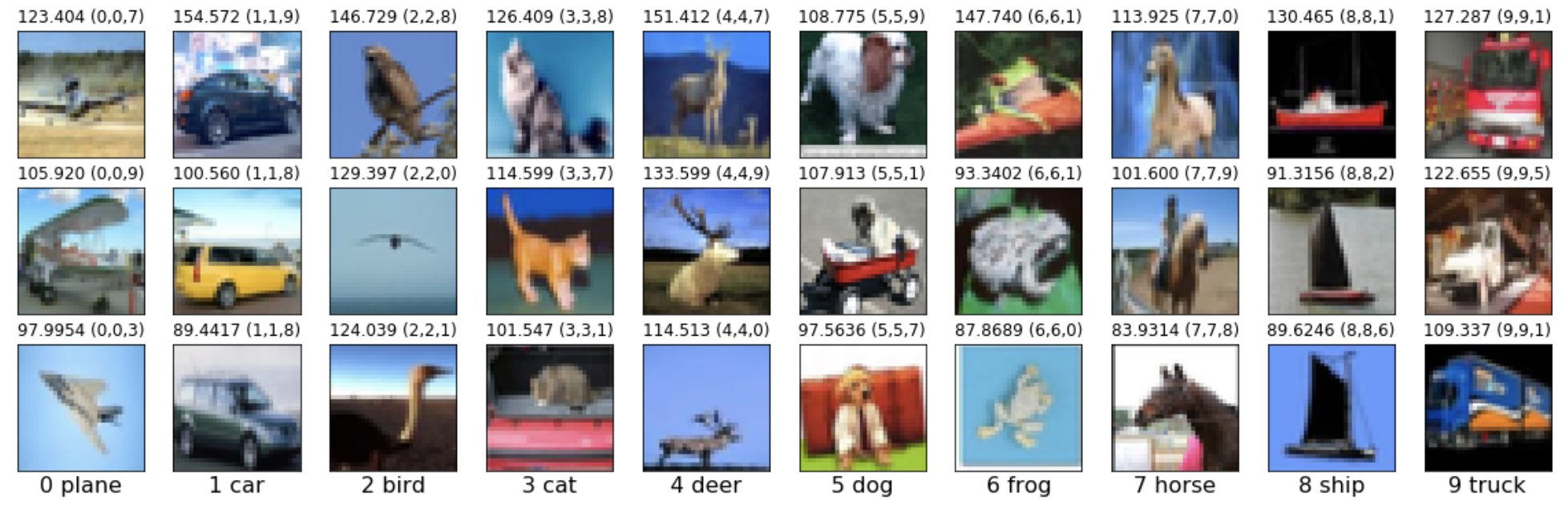}}%
	\caption{Low and high memorization score examples on CIFAR-10 dataset.
		Each column corresponds to one class.
		The number above the image represents the memorization score and the triplet above the image represents (ground truth label, unmuted sub-network's prediction, muted sub-network's prediction).}
	\label{fig:cifar10 easy and difficult examples}
\end{figure}

Figure \ref{fig:mnist easy and difficult examples} and Figure~\ref{fig:cifar10 easy and difficult examples} show the easy (low memorization score) and difficult (high memorization score) examples that are identified with turn-over dropout on MNIST and CIFAR-10 respectively. The easy examples do follow a typical pattern and shares a lot of similarities, and both the unmuted sub-network and the muted sub-network agree with the ground truth. The difficult examples seem to have some peculiar traits.

\section{Early Stop} \label{eary_stop}

\begin{figure}[h]
	\centering
	\subfigure[\small MNIST.]{
		\includegraphics[width=0.475\textwidth]{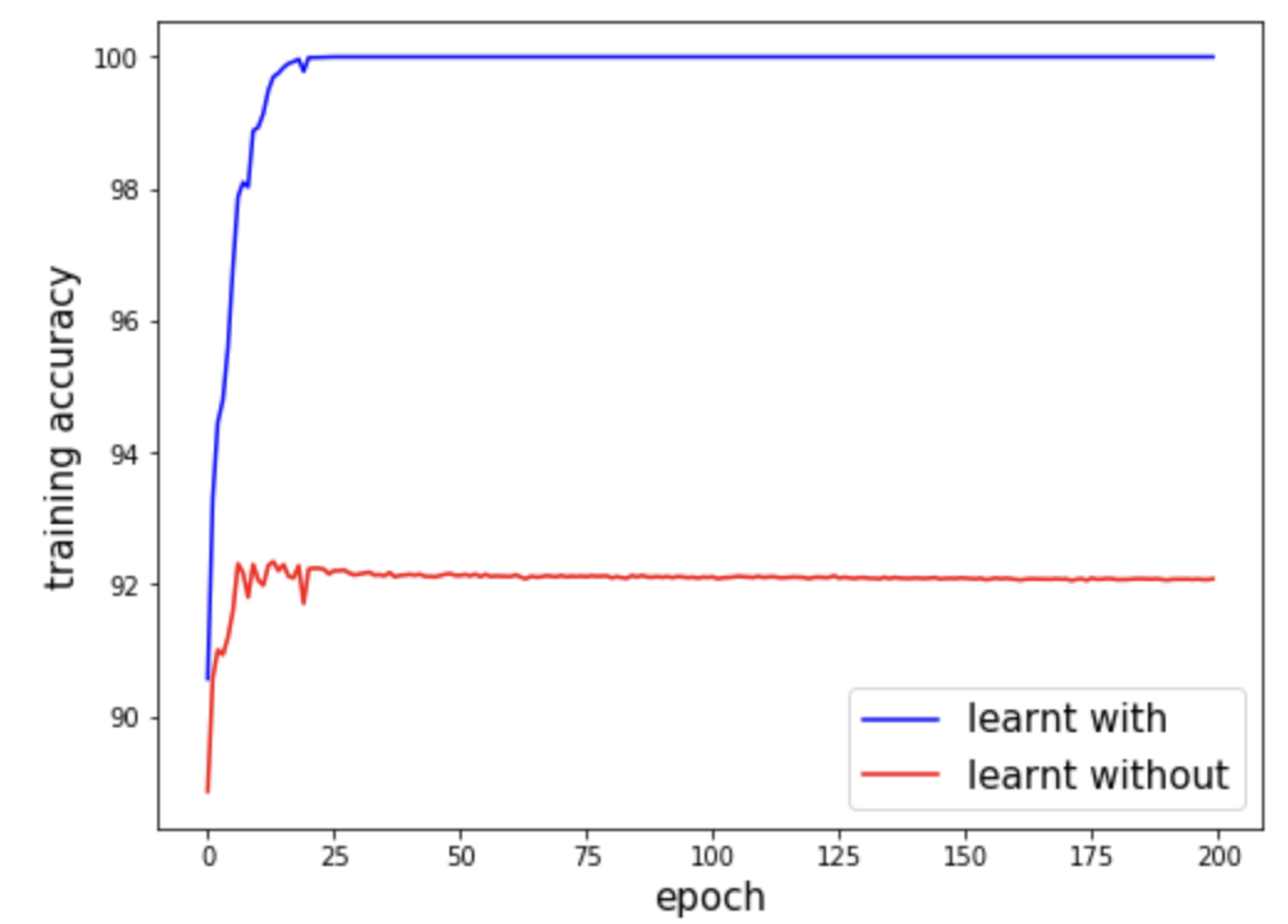}
	}
	\hfill
	\subfigure[\small Fashion MNIST.]{
		\includegraphics[width=0.475\textwidth]{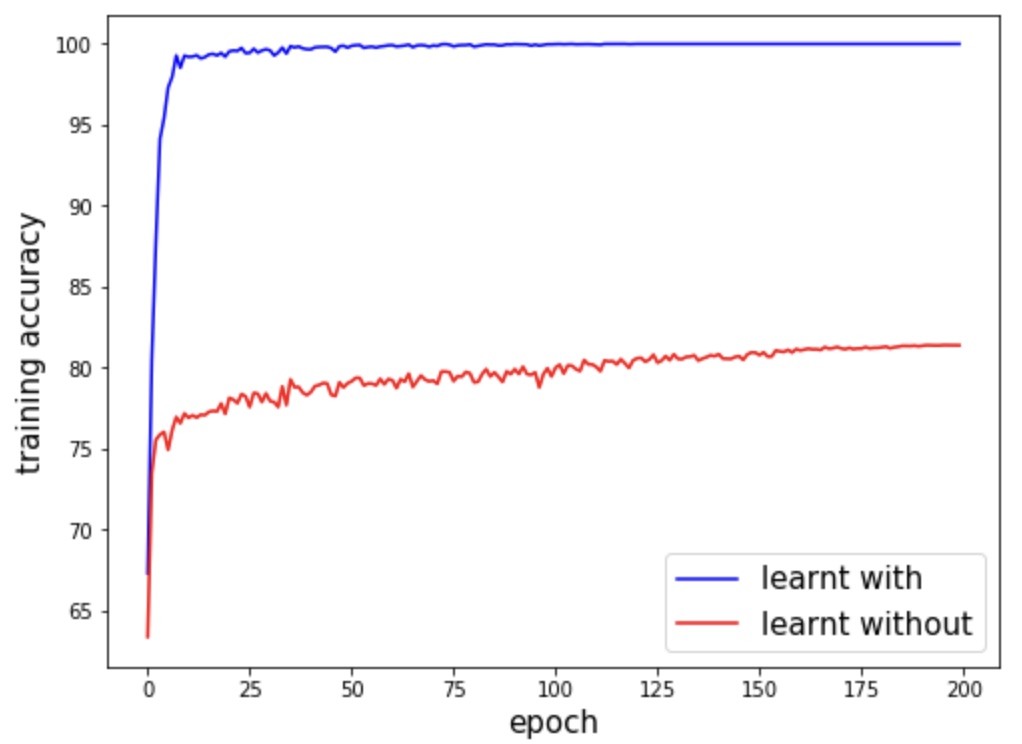}
	}
	\caption{\small
		\textbf{Training accuracy of the muted and unmuted sub-networks in a model with turn-over dropout}. (a) a one-hidden-layer MLP trained on MNIST with turn-over dropout. (b) a compact vision transformer \citep{hassani2021escaping} trained on fashion MNIST with turn-over dropout.
	}
	\label{fig:early stop with turnover dropout}
\end{figure}

Figure \ref{fig:early stop with turnover dropout} shows the training accuracy of the muted and unmuted sub-networks in a model with turn-over dropout. Both the muted and unmuted sub-networks converge very quickly within the first a few epochs and the stabilize. This motivates us to train the model with turn-over dropout for only a few epochs to calculate the memorization scores.

\section{Training and Test Dynamics of Different Degree of Corruption} \label{training_dynamics_on_random_labels}
\begin{figure}[ht]
	\centering
	\subfigure[\small Loss.]{
		\includegraphics[width=0.475\textwidth]{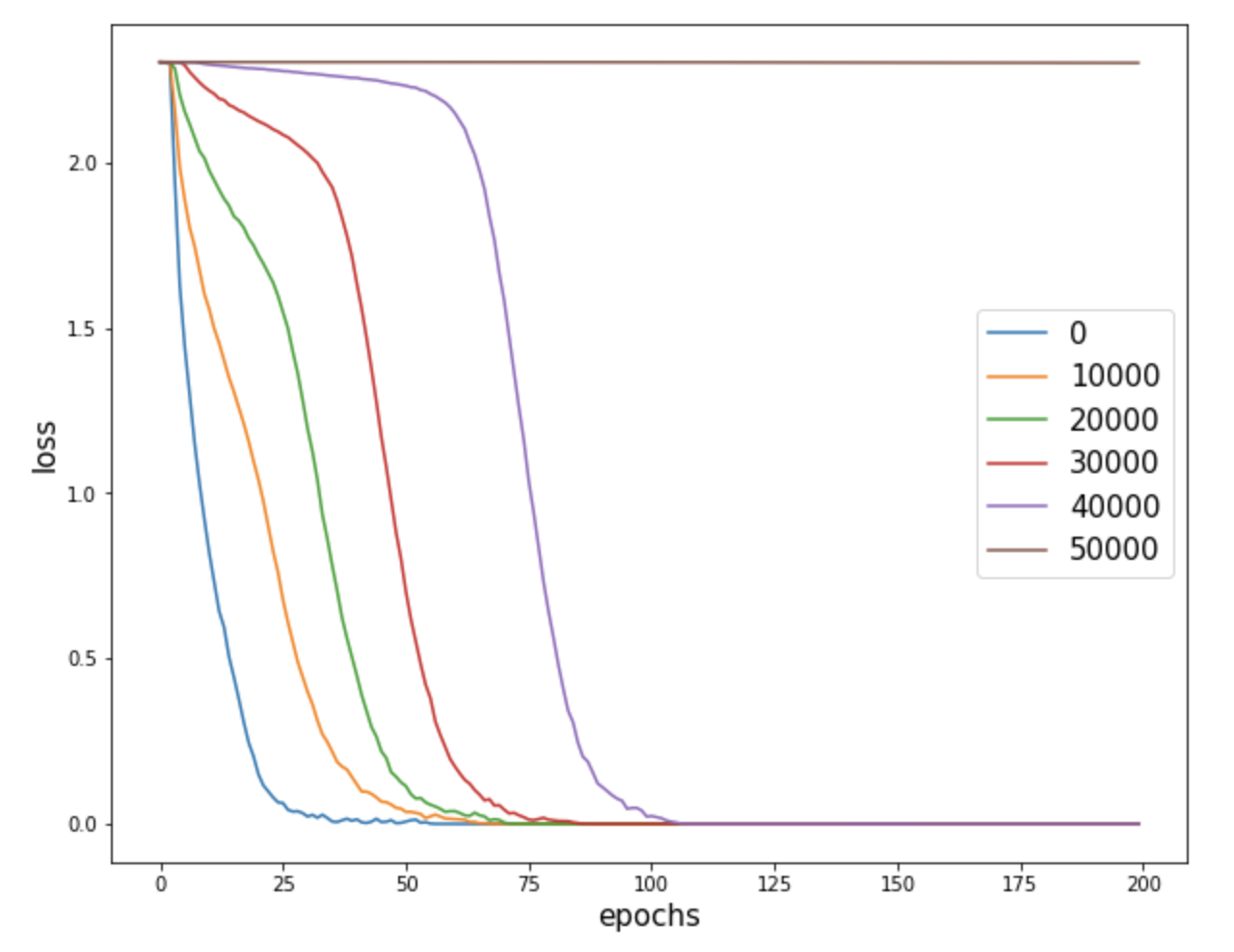}
	}
	\hfill
	\subfigure[\small Training accuracy.]{
		\includegraphics[width=0.475\textwidth]{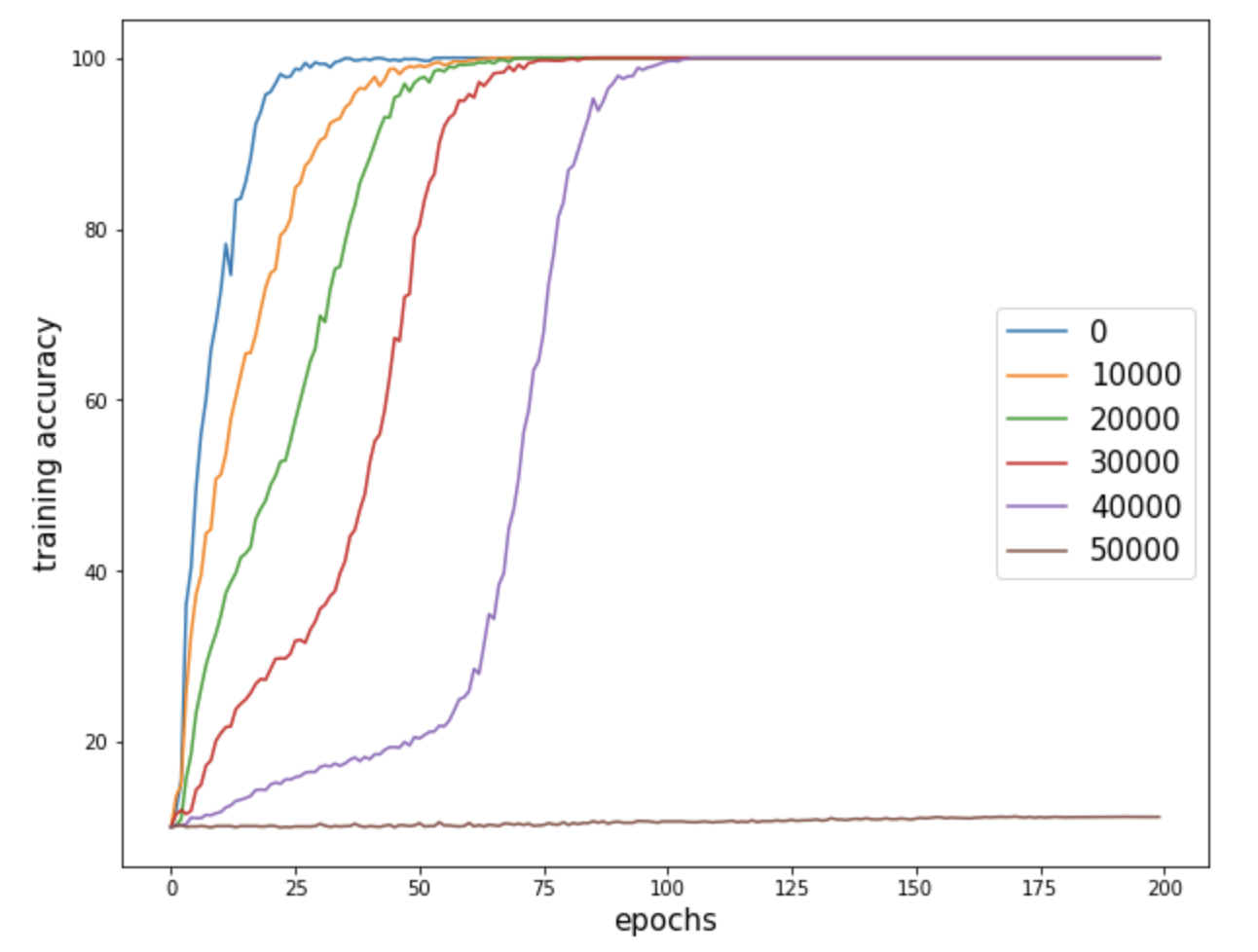}
	}
	\vfill
	\subfigure[\small Test accuracy.]{
		\includegraphics[width=0.475\textwidth]{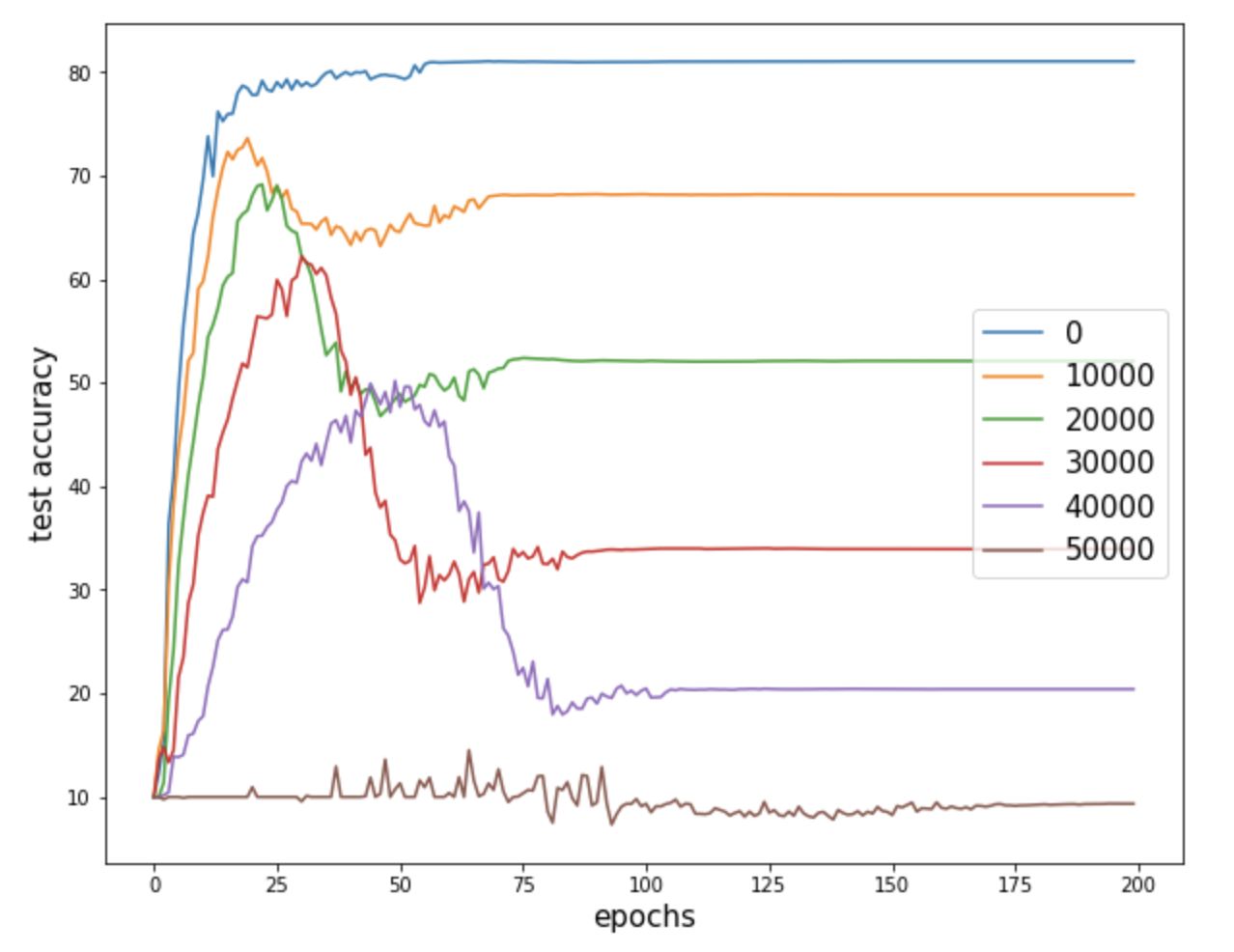}
	}
	\caption{\small
		\textbf{Training and test dynamics of VGG-11 network on CIFAR-10 where the training dataset is corrupted by different extent and the test dataset remains unchanged.}
		The legend shows the number of examples with random labels.
		It should be mentioned that since we use a multi-step learning rate decay schedule instead of a cosine annealing schedule, the model becomes incompetent to fit the dataset when all labels are replaced by random noise. This also proves that the extent and capability of memorization is influenced by the training procedure as well.
	}
	\label{fig:vgg on random labels}
\end{figure}

Figure \ref{fig:vgg on random labels} shows the training and test dynamics of a VGG-11 network trained on CIFAR-10, of which the labels of the training dataset are corrupted by different extent and the test dataset remains unchanged.

It can be observed that there are qualitative differences between the learning curves of different degree of corruption, which indicates that DNNs conduct memorization in a data-aware fashion.
There are two main observations:
\begin{enumerate}
	\item Difficult instances (random noises among the labels) make the learning process slower.
	      The more difficult examples, the longer it takes to fully fit the dataset.
	\item When there exists noise in the dataset, the test accuracy firstly reaches to a peak before it drops and converges.
\end{enumerate}

\end{document}